\documentclass[11pt]{article} 
\usepackage{rldmsubmit,palatino}
\usepackage{graphicx}
\usepackage{algorithmic}
\usepackage{amsmath,amssymb,amsfonts}

\title{Autonomous Open-Ended Learning of Interdependent Tasks}

\author{
Vieri Giuliano Santucci$^1$, Emilio Cartoni$^1$, Bruno Castro da Silva$^2$, Gianluca Baldassarre$^1$\\
$^1$Istituto di Scienze
e Tecnologie della Cognizione (ISTC)\\
Consiglio Nazionale delle Ricerche (CNR)\\
Roma , Italy\\
\texttt{\{vieri.santucci,emilio.cartoni,gianluca.baldassarre\}@istc.cnr.it}\\
$^2$Institute of Informatics\\
Federal University of Rio Grande do Sul (UFRGS)\\
Porto Alegre, Brazil\\
\texttt{bsilva@inf.ufrgs.br}\\
}

\begin{document}

\maketitle

\begin{abstract}
Autonomy is fundamental for artificial agents acting in complex real-world scenarios.
The acquisition of many different skills is pivotal to foster versatile autonomous behaviour and thus a main objective for robotics and machine learning. 
Intrinsic motivations have proven to properly generate a task-agnostic signal to drive the autonomous acquisition of multiple policies in settings requiring the learning of multiple tasks. 
However, in real-world scenarios tasks may be interdependent so that some of them may constitute the precondition for learning other ones. 
Despite different strategies have been used to tackle the acquisition of interdependent/hierarchical tasks, fully autonomous open-ended learning in these scenarios is still an open question.
Building on previous research within the framework of intrinsically-motivated open-ended learning, we propose an architecture for robot control that tackles this problem from the point of view of decision making, i.e. treating the selection of tasks as a Markov Decision Process where the system selects the policies to be trained in order to maximise its competence over all the tasks.
The system is then tested with a humanoid robot solving interdependent multiple reaching tasks.  
\end{abstract}


\acknowledgements{This project has received funding from the European
Union's Horizon 2020 Research and Innovation Program under
Grant Agreement no. 713010 (GOAL-Robots -- Goal-based
Open-ended Autonomous Learning Robots). This work was also partially supported by the Brazilian FAPERGS under grant no. 17/2551-000.}

\startmain 

\section{Introduction}

Autonomous acquisition of many different skills is necessary to foster behavioural versatility in artificial agents and robots. While the learning of multiple skills \textit{per se} can be addressed through different machine learning techniques sequentially assigning a series of $N$ tasks to the agent, autonomy implies that the agent itself has the capacity to select on which task to focus at each moment and to shift between them in a smart way. 
Intrinsic motivations (IMs) have been used in the field of machine learning and developmental robotics \cite{Oudeyer2007intrinsic, Baldassarre2013Book} to provide self-generated reinforcement signals driving exploration and skill learning \cite{Oudeyer2007intrinsic, Santucci2014,Hafez2017}. Other works \cite{Baranes2013,Santucci2016,Forestier2017} implemented IMs as a motivational signal for the autonomous selection of tasks (often called ``goals''): the learning progress in accomplishing the tasks is used as a transient reward to select goals in which the system is making the most learning progress \cite{Lopes2012,Santucci2013best}. 

In real-world scenarios, tasks may require specific initial conditions to be performed or may be interdependent, so that to achieve a task the agent needs first to learn and accomplish other tasks. This latter case is of particular interest and although it has been studied under different headings, it is still an open question from an autonomous open-ended learning perspective. 
Hierarchical reinforcement learning \cite{Barto2003} has been combined with IMs to allow for the autonomous formation of skills sequences. These methods often tackle only discrete state and actions domains \cite{Vigorito2010}; or focus on the discovery of sub-goals on the basis of externally-given tasks \cite{Bakker2004} or under the assumption that sub-goals come as predefined rewards \cite{Niel2018}, thereby reducing the autonomy of the agent learning process.
Imitation learning methods have also achieved important results in learning task hierarchies \cite{Grollman2010,Mohseni2018,Nair2018}, also in association with IMs \cite{Duminy2018}, but by definition they rely on external knowledge sources (e.g., an ``instructor''), which limits the agent's autonomy.

We propose a reinforcement learning (RL) system for robot control that is capable of learning multiple interdependent tasks by treating the selection of tasks/goals as a Markov Decision Process (MDP) where the agent selects goals to maximise its overall competence.

\section{Problem Analysis and Suggested Solution}

From an RL perspective, learning of multiple goals can be treated as learning different policies $\pi_g$, each one associated with a different goal state $g \in G$. Such policies aim at maximising the return provided by a reward function $R_g$ associated with goal $g$ (see e.g. \cite{Florensa2018}).
For each $g$ the system thus learn a policy
\begin{equation}\label{eqn:objFuncMDP}
   \pi_g^*(a|s) = \underset{\pi}{\text{argmax}} \, R_g(\pi_g)
\end{equation}
Since we are considering an open-ended learning scenario where no specific tasks are assigned to the robot, we assume that the system does not aim at maximising extrinsic rewards, but rather a competence function $C$ over the distribution of goals $G$. Here, $C$ is the sum of the agent's competence $C_g$ at each goal $g$ as made possible by a given candidate goal-selection policy $\Pi_t$. In other words, competence is a measure of the agent's ability to efficiently accomplish different goals by allocating its time among them using a given policy $\Pi_t$. Each goal can thus be associated with an MDP where the agent is tasked with learning to maximise the competence $C_g$ for that goal rather than the goal's extrinsic reward $R_g$. If we consider a finite time horizon $T$, the robot needs to properly allocate its training time to the goals that guarantee the highest competence gain.
To do so, the system may use the current derivative of the competence $\delta C$ (w.r.t. time) as an intrinsic motivation signal to select the goal with the highest competence improvement at each time step $t$, where time here refers to one training step over a given task. The problem of task selection can thus be described as an $N$-armed bandit (possibly a \textit{rotting bandit} due to the non-stationary transient nature of IMs) where the agent learns a policy $\Pi$ to select goals that maximise the current competence improvement $\delta C$:
\begin{equation}\label{eqn:objFuncBandit}
   \Pi^* =  \underset{\Pi}{\text{argmax}} \ \delta C (\Pi_t)
\end{equation}
The efficacy of this approach has been demonstrated in different works within the intrinsically motivated open-ended learning framework \cite{Lopes2012,Merrick2012,Santucci2013iccm}. If we constrain the feasibility of the goals to specific environmental conditions, goal selection becomes a \textit{contextual bandit} problem where the robot has to learn to select goals depending on its current state $s \in S$. Equation \ref{eqn:objFuncBandit} thus becomes:
\begin{equation}\label{eqn:objFuncContestualBandit}
   \Pi^*(s_t) =  \underset{\Pi}{\text{argmax}} \ \delta C (\Pi(s_t))
\end{equation}
\noindent where now the policy for selecting the goals to train needs to explicitly take into account the current state of the agent, which may include information such as which other goals have already been accomplished. By making this change to the objective, the system can bias the choice using the expected competence gain for each goal given different conditions. The evaluation of the competence improvement for each goal can be done via a state-base moving average of performance at achieving that goal given the current policy. 
If we now further assume a situation where goals are \textit{interdependent}, so that a goal may be a precondition for other ones, we shift to a different kind of problem where the state of the environment depends on previously selected (and possibly achieved) goals. A sequence of contextual bandits where the context at time $t+1$ is determined by the ``action'' (here, goal selection) executed at time $t$, can be seen as an MDP over all the goal-specific MDPs for which the robot is learning the policy (a ``skill''). This is a setting that hierarchical skill learning methods have only scarcely addressed within a fully autonomous open-ended framework.

In this paper we propose that, given the structure of the problem, goal selection in the case of multiple interdependent tasks can be treated as an MDP and, consequently, can be addressed via RL algorithms that transfer \textit{intrinsic-motivation} values between interrelated goals. 
In particular, in the following sections we show how a system implementing goal selection with a standard Q-learning algorithm is able to outperform systems that treat goal selection as a standard bandit or contextual bandit problem.

\section{Experimental Setup and System Comparison}
\label{sec:M-GRAIL}

\begin{figure}[ht!]
    \begin{center}
    \includegraphics[width=.20 \textwidth,keepaspectratio]{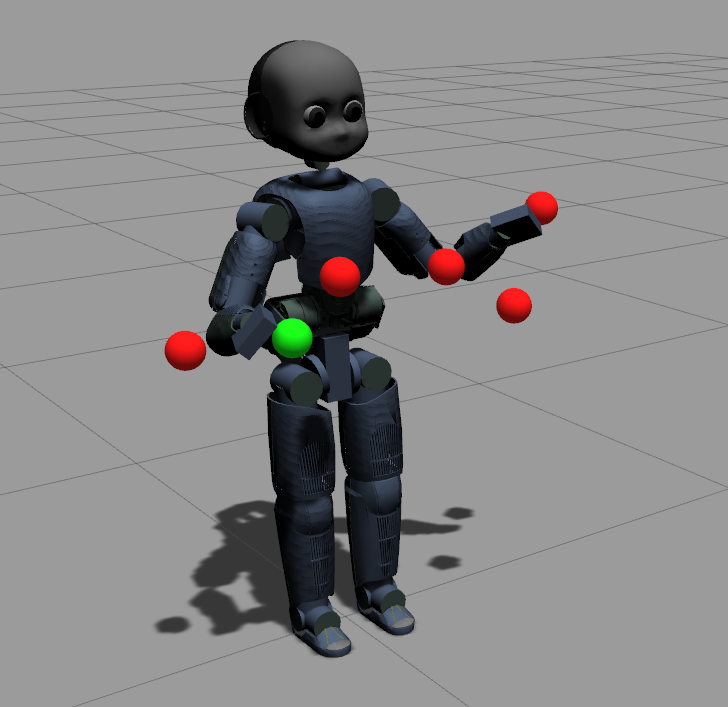}
    \end{center}
    \caption{The simulated iCub robot in our experimental setup: when a sphere is touched (given its preconditions) it ``lights up'', changing its colour to green.}
    \label{fig:Robot}
\end{figure}

To test our hypothesis we compared different goal-selection systems in a robotic scenario with a simulated iCub robot (Fig.~\ref{fig:Robot}) that has to reach and ``activate'' 6 different spheres. We present results comparing three algorithms (discussed below) in two experimental scenarios: \\
    \textbf{1)} \textit{Environmental Dependency/Contextual Bandit Setting}: the activation of a sphere, by having the robot touch it, is dependent on some environmental variable. In this setting we assume a state feature (the ``contextual feature'') that is set to 1.0 with 50\% probability at the beginning of each trial, and to 0.0 otherwise. The environment is composed of a total of six spheres that the agent needs to learn to activate: three can only be activated when the contextual feature is on, and the other three only when it is off.\\
    \textbf{2)} \textit{Multiple Interrelated Tasks/MDP Setting}: the ``achievability'' of a task (activation of a sphere) is now dependent on the activation status of the other spheres. In this scenario, the fact that the robot has previously achieved or not a goal (or set of goals) constitutes the precondition for the achievement of other goals, thus introducing interdependencies between the available tasks.
%
\begin{figure}[ht!]
  \begin{center}
    \begin{minipage}{.35\textwidth}
       \centering
       \includegraphics[width=1\columnwidth]{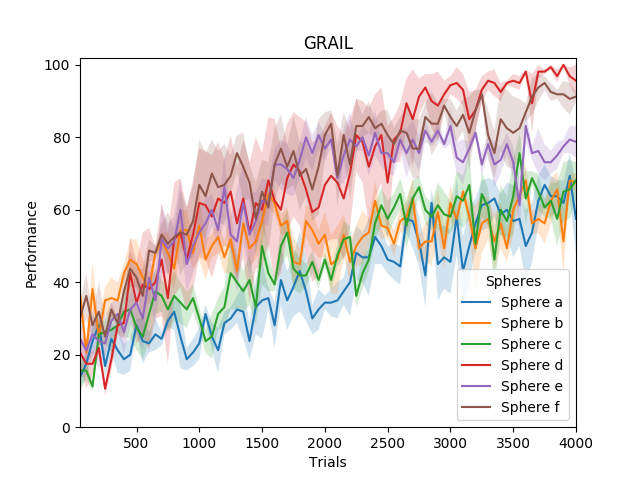}
    \end{minipage}
    \begin{minipage}{.35\textwidth}
       \centering
       \includegraphics[width=1\columnwidth]{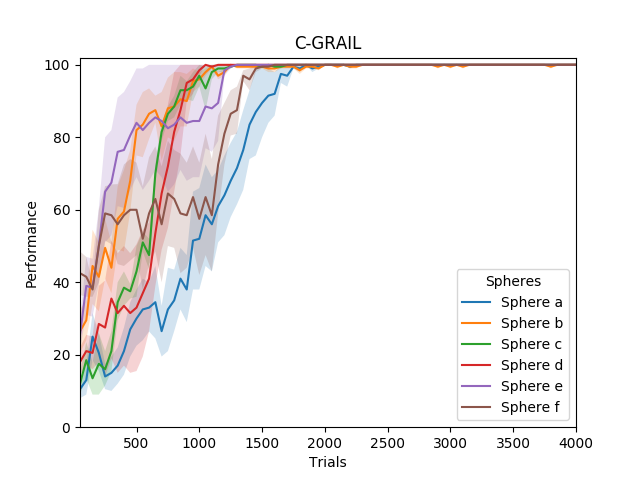}
    \end{minipage}
  \end{center}
    \caption{Performance of GRAIL and C-GRAIL in the first experiment}
    \label{fig:Exp2}
\end{figure}

In this paper we compare three goal-selection systems that build upon the existing GRAIL architecture \cite{Santucci2016}. This architecture is generally composed of two components: a high-level component, called the ``goal selector'' (GS), which performs task selection on the basis of competence-based intrinsic motivations; and a low-level component composed of a set of low-level experts (one per task or goal); each expert is an actor-critic neural network implementing a candidate policy for achieving a goal.
In the original version of GRAIL, the GS component receives no input from the environment and selects goals as in a standard bandit setting, where each arm/goal is evaluated on the basis of an exponential moving average (EMA) of the previously-acquired intrinsic rewards associated with achieving that goal. We now discuss two different versions of GRAIL that, by modifying the GS component, are able to cope with the added complexity of the scenarios described above. The first, called \textit{Contextual}-GRAIL (C-GRAIL), provides as input to the GS the state of the environment, which can be composed of standard state features or also features describing the status of different goals (e.g features describing whether each sphere is activated). The GS then selects tasks to practice as in a contextual bandit where different EMAs are associated with different contexts. A second possible modification to GRAIL, called \textit{Markovian}-GRAIL (M-GRAIL), provides the same input to the GS as in C-GRAIL, but treats goal selection as an MDP and solves it by modeling the temporal interdependency between goals as the temporal dependency between consecutive states in an MDP; it then uses Q-learning to assign a value to each goal, where values represent the long-term benefits of practicing that goal considering the intrinsic rewards that goals that depend on it may provide in the future.

\section{Results}

In our first experiment we compare GRAIL and C-GRAIL in a setup where the value of a contextual feature is used as precondition to determine whether the agent can activate certain spheres. In particular, spheres $a$, $c$ and $e$ can only be activated when the contextual feature ($cf$) is set to 1.0, while spheres $b$, $d$ and $f$ can only be activated when $cs$ is set to 0.0. At the beginning of each trial, $cf$ is set to 1.0 with 50\% probability. While GRAIL selects tasks without considering the environmental condition, C-GRAIL receives $cs$ as input and performs task selection as in a contextual bandit. Fig.~\ref{fig:Exp2} shows the performances of GRAIL and C-GRAIL on the 6 tasks during an experiment that lasts for 4000 trials. At the end of each trial, the environment is reset (all spheres are set to ``off''). C-GRAIL is able to properly learn all tasks in approximately 2000 trials, while GRAIL (at the end of the simulation) has achieved high competence in only two of the tasks. This is because GRAIL performs selection (and value assignment) without taking the status of the $cs$ into account, which is by construction important to determine whether spheres can be activated. While C-GRAIL can properly generate IMs for the different tasks only in those conditions where they can be in fact be achieved, GRAIL ``wastes time'' in selecting tasks even when they cannot be trained, thus impairing the learning process.
\begin{figure}
  \begin{center}
    \begin{minipage}{.35\textwidth}
       \centering
       \includegraphics[width=1\textwidth]{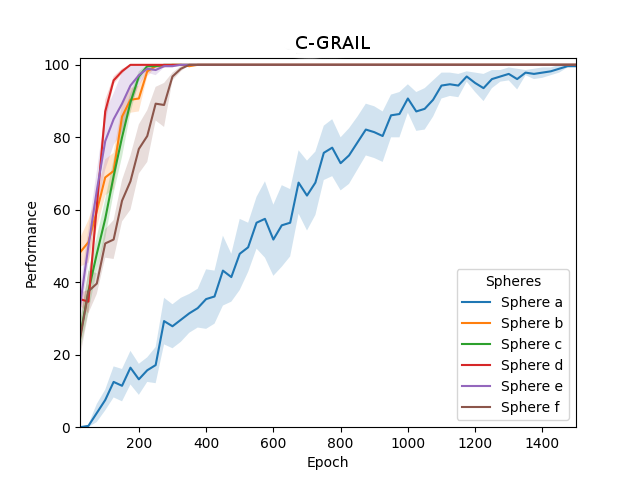}
    \end{minipage}
    \begin{minipage}{.35\textwidth}
       \centering
       \includegraphics[width=1\textwidth]{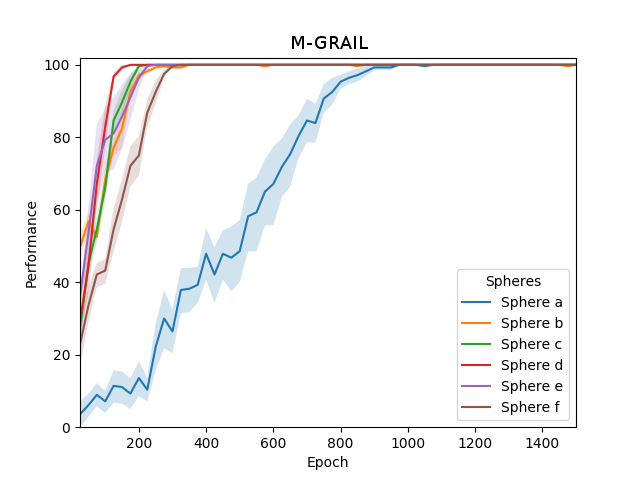}
    \end{minipage}
  \end{center}
    \caption{Performance of C-GRAIL and M-GRAIL over the 6 goals in the second experiment}
    \label{fig:LWorld5_Performance_WastedTrials}
\end{figure}

In our second experiment we introduce interdependencies between goals. In particular, sphere $f$ can be activated only if sphere $c$ is already active, while sphere $a$ can be activated only when $f$ is on. This implies that to light up sphere $a$, the robot has to first turn on spheres $c$ and $f$. Other spheres have no dependencies. We ran simulations for 1500 epochs, each one lasting 4 trials, for a total of 6000 trials. At the end of each epoch we reset the environment; during each epoch, spheres retain their current status as determined by the actions of the robot.

Based on the first experiment we can observe that GRAIL is not capable of properly performing autonomous learning when tasks are dependent on preconditions; we thus evaluated C-GRAIL and M-GRAIL to study whether they help tackle the interdependent-task setting. Both algorithms provide as input to the goal selector, at each step, the status of the six spheres (``on'' or ``off''); however, they assign values to each candidate goal in different ways, as briefly discussed in Section \ref{sec:M-GRAIL}. By comparing the performances of C-GRAIL and M-GRAIL (Fig.~\ref{fig:LWorld5_Performance_WastedTrials}) we observe that while M-GRAIL reaches a perfect overall performance after about 975 epochs, C-GRAIL achieves a high performance on all goals only at the very end of the experiment.

Although both systems are capable of assigning positive values to goals only in states where their preconditions are satisfied, C-GRAIL is negatively affected whenever achieving a task such as $a$ requires that the agent first satisfy a number of previous preconditions. Intuitively, $a$ is ``distant'' from the initial condition of the system, where all spheres are off. Furthermore, whenever a task is completely learned (the robot has an optimal policy for performing a goal), the intrinsic motivation for selecting it gradually disappears. This may lead to a situation where the robot starts selecting tasks almost at random due to the absence of intrinsic rewards, thus wasting trials in selecting goals that cannot be achieved at that moment. As a result, even though the robot may have an intrinsic motivation reward for practicing a goal (e.g. activating sphere $a$), it does not have intrinsic motivation for first practicing the goals that are preconditions to $a$; it is not, thus, capable of systematically putting the environment in the proper conditions to train $a$. M-GRAIL, by contrast, can rapidly learn all tasks: even when (similarly to C-GRAIL) it is no longer intrinsically motivated in achieving ``simple'' goals \textit{per se} (i.e., goals with few preconditions), it ensures that the robot continues to select those goals thanks to the Q-learning algorithm, which propagates the intrinsic motivations for solving task $a$ back to the tasks that are $a$'s preconditions.

\bibliographystyle{abbrv}
\bibliography{RLDMBib}

\end{document}